# When is it right and good for an intelligent autonomous vehicle to take over control (and hand it back)?

Ajit Narayanan

Department of Computer Science

School of Engineering, Computer and Mathematical Sciences

Auckland University of Technology

Auckland

New Zealand

## Abstract

There is much debate in machine ethics about the most appropriate way to introduce ethical reasoning capabilities into intelligent autonomous machines. Recent incidents involving autonomous vehicles in which humans have been killed or injured have raised questions about how we ensure that such vehicles have an ethical dimension to their behaviour and are therefore trustworthy. The main problem is that hardwiring such machines with rules not to cause harm or damage is not consistent with the notion of autonomy and intelligence. Also, such ethical hardwiring does not leave intelligent autonomous machines with any course of action if they encounter situations or dilemmas for which they are not programmed or where some harm is caused no matter what course of action is taken. Teaching machines so that they learn ethics may also be problematic given recent findings in machine learning that machines pick up the prejudices and biases embedded in their learning algorithms or data.

This paper describes a fuzzy reasoning approach to machine ethics. The paper shows how it is possible for an ethics architecture to reason when taking over from a human driver is morally justified. The design behind such an ethical reasoner is also applied to an ethical dilemma resolution case. One major advantage of the approach is that the ethical reasoner can generate its own data for learning moral rules (hence, 'autometric') and thereby reduce the possibility of picking up human biases and prejudices.

The results show that a new type of metric-based ethics appropriate for autonomous intelligent machines is feasible and that our current concept of ethical reasoning being largely qualitative in nature may need revising if want to construct future autonomous machines that have an ethical dimension to their reasoning so that they become moral machines.

Keywords: Moral machines; artificial morality; intelligent autonomous vehicles; autonomous intelligent systems.

## 1. Background and introduction

The International Federation of Automatic Control defines "intelligent autonomous vehicles" (IAVs) as "automated vehicles capable of performing motion control tasks in unstructured or partially structured environments with little (if any) assistance from human supervisors."[1] The potential capability of IAVs was still in doubt as recently as 10 years ago, with research in IAVs confined to simple laboratory environments due to difficulties of finding ways to incorporate and integrate intelligent sensing, reasoning, action, learning and collaboration [1]. The most promising developments up to that time included subsumption architectures, where IAV perception was linked to action without the need for an internal representation of the environment, and with multiple control layers where each represented a behaviour [2,3]. Subsumption architectures led to major developments in walking robots [4], small rovers for Mars missions [5] and sociable robots [6]. A more classical, symbolic control approach was Autonomous Robotic Architecture (AuRA), where different modules for planning, reasoning and motion interacted via schemas for collision avoidance and problem resolution [7]. Perhaps the most successful approach was the Intelligent Controller (IC) architecture, which combines the subsumption approach with modules to create internal representations from incoming sensor data to fuse with previous sensor data [8, 9]. The IC architecture has been successfully applied to several unmanned aerial and underwater vehicles (e.g. [10]). Other developments included enhancing general cognitive architectures such as Soar [11] and ACT-R [12], with perception and actuation modules for interaction with the environment. Such cognitive architectures are characterized by the use of production rules and supplemented with additional algorithms for dealing reactively [13,14]. Real-time control system (RCS) developed at NIST is a reference model architecture involving a systematic mapping using nodes, where a node consists of behaviour generation, sensory processing, world modelling and evaluation. RCS was applied to autonomous on-road driving in 2004 [15]. Underlying all IAV approaches so far are three basic stages: sensing and processing, decision making, and reacting.

DARPA's sponsorship of a series of AV challenges started in 2004, when a competition was held to self-navigate over 140 miles of desert roadway in 10 hours. The aim at that time was that a third of military vehicles should be AVs by 2015. But in 2004 no AVs could proceed more than a few miles without crashing. Improvements in control software, collision avoidance, road following as well as radar and laser sensing technologies contributed to significant advances so that, in 2007, the route was changed to 60 miles of urban conditions. Four AVs completed the task in the six-hour time limit allowed. Since that time, advances have led to several spin-offs in conventional vehicles, such as lane adhesion, emergency braking and self-parking. Currently, AV technology and development are being driven by major car manufacturers such as Mercedes, Nissan, BMW, VW, Volvo and GM, with Google emerging as another major contributor in 2009 with its Self-Driving Car Project (evolved to Waymo in 2016) building on its Google Maps data to recognize locations. In 2015 Tesla introduced the Model S which has autonomous steering, side collision avoidance, lane changing and parallel parking capabilities. Software updates to its Autopilot system now allow Model S to self-park without the driver being in the car. Also in 2015 Delphi Automotive developed an AV that drove over 3000 miles coast-to-coast across North America under autonomous control for 99% of the distance.

---

[1] https://tc.ifac-control.org/7/5/activities/ifac-intelligent-autonomous-vehicles-iav-symposium

A classification system was released in 2014 by the automotive standardization body SAE International, based on the amount of driver attention required [16]. Level 0 is driver only, where the driver manages all driving aspects (steering, speed, monitoring of driving environment). Level 1 is assisted, where the driver is given support for either steering (e.g. parking) or speed (e.g. cruise control) in specific situations. Level 2 is partial automation, where the driver is given support for both steering and speed (e.g. lane adherence with cruise control) but must continue monitoring the driving environment to intervene when necessary. Level 3 is conditional automation where the driver can relinquish control to an automated driving system which controls steering, speed and monitoring of the driving environment but must be ready to take back control. Level 4 is high automation where the automated driving system controls all aspects of the dynamic driving task (steering, braking, speed, environment monitoring, changing lanes) even if a human driver does not respond appropriately to a request to intervene. Finally, Level 5 is complete end-to-end journey without any driver intervention. The distinctions between "automated", "automatic" and "autonomous" are not always clear [17, 18], but it is generally accepted that autonomous vehicles are characterized by vehicles achieving levels 3, 4 and 5 of the SAE classification system, where the driver relinquishes control of the driving environment either partly or fully.

The need for artificial intelligence and intelligent decision-making to play a role in autonomous control systems was recognized early in the 1990s [17]. Since that time, major advances in autonomous vehicles have been driven by increases in processing power and big data. Supercomputers can now process massive amounts of sensor, camera and radar data in processors capable of operating at over three hundred trillion operations per second for the sensing and processing stage. Associated with these technological advances is "deep learning", where training data is fed into neural networks consisting of dozens or even hundreds of layers on GPU-accelerated platforms [19] for learning about traffic conditions and making decisions. But alongside predictions by industry commentators that advances in big data and deep learning are leading to level 5 IAVs within a matter of years [20], there are signs of scepticism that full autonomy will be achieved in such a short time as well as worries that over-expectation can lead to an 'AI winter' [21]. The main problem appears to be that, despite technological advances in sensors, architectures and processing power, it is not clear how to program basic "common sense" for dealing with new situations into such technologies [22]. The main reason for this scepticism lies in the nature of the accidents involving autonomous vehicles that have occurred up to now.

One of the main motivations for autonomous vehicles is that they are intended to be safer than vehicles controlled by the average human driver. For 2015, there were approximately 3.6 road fatalities per 1 billion vehicle-driven km in the UK, 7.1 in the USA and 8.7 in New Zealand [23]. Autonomous vehicles and especially fully-autonomous vehicles have not been driven for long enough for comparable fatality numbers to be reliably calculated. So far, there have been four fatalities involving partially autonomous vehicles: three involving Tesla Autopilot (twice in 2016, once in 2018) and one involving Uber (in 2018).

In January 2016, a Tesla car in Hebei, China, crashed into the back of a road-sweeping truck, killing the driver. The lack of any evidence of car braking or swerving has led to claims that the Autopilot was engaged but failed to work properly. In May of that year, a Tesla car in Williston, Florida, crashed into a tractor trailer while in Autopilot, with the cause identified as the white side of the trailer not being distinguished from the brightly lit sky and so the brake not applied. In March 2018, a self-driving Uber car killed a pedestrian in Tempe, Arizona, when she walked her bicycle across a street at 10pm. Also in March 2018, in Mountain View, California, a Tesla car in Autopilot set for 75 miles an hour crashed into a safety barrier, killing the driver and causing two other vehicles to crash into it. There is still uncertainty as why the vehicle hit the barrier and why the driver did not take avoidance action in response to warning sounds. Such fatalities inevitably raise questions concerning the need for improved sensor technology as well as enhancements to both the decision making and action stages of control.

However, one critical aspect is frequently ignored when discussing the role of AI in autonomous control systems, which is the need to ensure that autonomous cars realise that it is wrong to take actions (or not to take actions) that can lead to humans being killed. In other words, while technological advances in sensor technology, architectures, hardware and programming may well lead to improvement in autonomous vehicle safety, another approach is to ensure that autonomous vehicles under intelligent control acquire an ethical sense that will ensure that their sensor-based decisions do not harm humans no matter what control architectures or types of technology are being used. This, after all, is what underlies human driver behaviour irrespective of the actual car being driven or driver-assisted technology being used in that car.

It is important to distinguish three types of relationship between ethics and autonomous systems. The first type is ethical design, which is a method for encouraging the design of systems for human values [24] and the consideration of ethical issues when designing and developing systems [25]. The IEEE Standards Association has recently launched a global initiative on ethical design approaches for autonomous and intelligent systems which outlines ethically aligned design around the principles of human rights, well-being, accountability, transparency and awareness of misuse [26]. A number of content committees and working groups are currently working on recommendations for standards to ensure that ethical considerations are prioritized in design and development of autonomous systems for the benefit of humanity. The second type is the ethics of autonomous systems and the consideration of whether it is right or wrong to construct such systems [27]. Such considerations take in to account the possibly dehumanizing aspects of research into AI as well as implications, such as loss of jobs and the dangers of superintelligence [28].

The third type is machine ethics, which is the area that concerns how we give ethical principles to intelligent systems so that such systems can decide for themselves what is right and wrong [29]. The problem is that programming ethical principles into an autonomous system is like hard-wiring the system so that it must follow these principles. The contradiction here is that such hard-wiring goes against the notion of intelligent autonomous systems that are supposed to make informed decisions for themselves [30]. "I avoided killing the pedestrian because I'm hard-wired to do so" is a deontological statement that does not allow for exceptions, as might be the case when the vehicle has to decide whether to run over ten children crossing the road or one person on the pavement with no other options available. For an autonomous system to do only what it is told to do raises questions as to what autonomy means. More importantly, it raises questions as to whether the four fatalities that have so far occurred were due to the system not being able to do other that what it was programmed to do. In other words, because there was no ethical component in the autonomous system, it followed instructions and executed actions blindly with no concept of the harm that such actions could cause.

The overall purpose of this paper is to explore the possibility of introducing an ethics reasoner into intelligent control systems in general, and IAVs in particular. While there is much research and debate concerning the rights and wrongs of AI, as well as what forms of ethics should be built into machines, the issue of how we design and implement an ethical reasoner for use in intelligent, autonomous machines is relatively unexplored. The purpose of this paper is to explore how ethical decision-making can be built into autonomous control systems so that future IAVs have an ethical dimension to their behaviour. More precisely, the aim is to explore methods for making autonomous vehicles calculate and behave as moral agents [31] and so shed light on 'artificial morality' [32,33].

Three problems need to be addressed [34] when designing, developing and implementing an ethical or moral machine. The first problem concerns the type and degree of *interactivity* that allows the moral machine to respond ethically to its environment. The second is the degree of *autonomy* from the environment that allows the moral machine to go through an ethical reasoning process of its own. And the third is amount of *adaptability* the moral machine is allowed to change its ethical reasoning processes. Together, these three desirable properties provide the basis for a *trustworthy* moral machine. The extent of the trust placed on such moral machines will depend on how it responds to different ethical situations and its ability to provide justifications for its responses.

We show how a fuzzy logic approach, which has the ability to reason under inexact or partial sensor knowledge, can produce a spectrum of ethical outputs based on different types and degree of interactivity. We will demonstrate through simulation and experiment how autonomous ethical decision making can be undertaken separately from environment sensor data. Finally, we will show how the decision-making system can adapt to new and changing situations, especially situations containing dilemmas. As far as we are aware, this is the first time that a fuzzy logic approach has been used to demonstrate ethical decision making for possible use in a moral machine.

## 2. Previous work

Ethical theories deal with rules or criteria for distinguishing right from wrong as well as good from bad. Examples of ethical theories are deontology (we must act according to duties and obligations), categorical imperative (we must act in accordance with human rational capacity and certain inviolable laws), utilitarianism (an action is right or good if it leads to most happiness) and consequentialism theories in general (whether an action is right or good depends on the action's outcome or result). Another approach is virtue ethics (we must act in ways that exhibit our virtuous character traits), where a trait is what allows us to fulfil our function. For humans, one specific function is to think rationally, and so virtue ethics is ethics led by reason to perform virtuous actions.

Previous approaches to implementing ethical reasoning in computers have not always clearly identified the ethical approach adopted and have involved case-based reasoning, neural networks, constraint satisfaction, category theory, and abductive logic programming as well as inductive logic programming. We present a brief overview below.

Early attempts in case-based reasoning approaches include Truth-Teller and SIROCCO, where the former identified shared features in a pair of ethical dilemmas and the latter retrieved ethical cases similar to a new case [35]. However, case-based reasoning systems in general are intended to support human ethical decision-making rather than help machines perform ethical reasoning on their own. Artificial neural network (ANN) approaches that learn ethical outputs from training samples [36] require specific topology and learning architectures for successful testing, with uncertainty concerning how to characterize the type of moral reasoning involved in the learning interaction. Also, the lack of rule-based reasoning capability internally or as output can lead to criticisms that such networks lack both transparency and autonomy. That is, such networks can only go through an internal and possibly ethically uninterpretable transition when given an input from the environment. Constraint satisfaction approaches [37, 38], while useful for certain types of AI problems requiring optimal solutions that do not violate conditions, assume full observability of the world ('closed world assumption') that can cause problems when knowledge is partial, vague or uncertain. Also, the need not to violate constraints is a form of deontology: actions are right or wrong depending only on rules rather than consequences. This makes the application of constraint satisfaction approaches to ethical dilemmas difficult, since dilemmas involve a decision to be made between two opposing constraints. Category theory approaches [39, 40] lead to ethical reasoning being interpreted as a functional process of mappings, or morphisms, from a domain of entities to a codomain. The use of category theory in machine ethics applies this formal approach so that an ethically relevant decision is correct if a formula containing that decision can be identified and proved as a theorem in an axiomatic system. As noted earlier, formal rule-based approaches to machine ethics, such as category theory and constraint satisfaction, raise questions concerning genuine autonomy. Such machines can only do what they are programmed to do within the formal system. Prospective, or abductive, logic approaches [41] attempt to 'look ahead' to future states before selecting *a posteriori* preferences. While the application of such an approach in machine ethics has the advantage of allowing a degree of consequentialism, the choice between preferences needs the support of a knowledge base and a set of non-violable integrity constraints. Such abductive logic approaches, similar to the other formal approaches of constraint satisfaction and category theory, depend on the closed world assumption of a preference not being against known principles and constraints which are hardwired into the program. Finally, inductive logic programming approaches can be used for machine learning of *prima facie* duty theory, and where there is no single absolute duty to be adhered to for deciding whether an IAV should or should not take over control [42,43]. The requirement is for a list of binary ethical features, a list of duties for minimizing or maximizing, and a list of actions. A number of cases can be represented in these data structures for machine learning of ethical principles, together with preferable actions as target (class) values, leading to a standard training-testing regime for learning when, morally, to take over control from a human driver. However, it is not clear how non-binary features can be handled (e.g. the varying desirability of respecting driver autonomy depending on continuously changing sensor information). Nor is it clear how missing, partial or inexact values affect the learning of ethical principles, since all features need to have values for the inductive engine to operate on a complete and consistent basis. Finally, there is a slowly emerging consensus that inductive machine learning algorithms with human specified feature values could be subject to algorithmic bias or learn biases in supplied data [44,45,46]. It is especially important for moral machines not to learn specific ethical preferences of programmers or biases in supplied data if they are to be considered trustworthy.

In summary, previous applied work in machine ethics does not address all three of the desirable properties of machine ethics. Interactivity is typically implemented through fixed data input (e.g. training data) rather than sensors that produce dynamically changing data. Previous approaches have not shown how moral decision making can vary through interaction with a dynamic environment. Adaptability is implemented as classifying unseen cases after successful training, as in the case of inductive logic programming and ANNs. However, another more intuitive sense of adaptability is the generalization or application from what is known from previous cases so that moral decisions continue to be made consistently for situations not previously encountered. Finally, there is a tendency to 'over-prescribe' the system with strict and formal moral rules, leading to questions as to how much genuine autonomy a machine ethics system contains. In particular, moral decision-making is typically based on formal rule-following rather than internal reasoning based on state-matching and state-transition, both of which may be approximate or imprecise.

As can be seen from the above, machine ethics is a comparatively under-explored area of artificial intelligence in general and machine learning in particular. The aim of this paper is to explore a radically different approach to machine ethics using inexact, or fuzzy, reasoning that aims to address the problem of ethical decision making. Any approach to machine ethics must also demonstrate that it can cope with ethical dilemmas, since such dilemmas test the ability of the system to go beyond what it has learned to do in specific situations to situations not previously encountered. In particular, dilemmas test the ability of an ethical system to balance conflicting aspects of duty against consequences.

## 3. Machine ethics design considerations

A characteristic of nearly all previous attempts to get machines to perform ethical reasoning is the use of a normative approach that can be called 'top down': machine ethics systems are designed and built with strict rules already in place for drawing conclusions. As noted above, this leads to questions concerning genuine autonomy of such systems. When rules are not explicitly given to the system, as in the case of ANNs, there are problems of possible data bias as well as lack of transparency concerning reasons for the moral output. A much better approach may be to provide a minimal set of principles rather than hard rules that can be universally agreed with and then allow the machine ethics system 'decide for itself' (autonomy) how to apply those principles in dynamically changing environments (interactivity) to derive moral rules that will allow it to monitor and change its behaviour in the light of new information (adaptability).

Another problem is that previous approaches, apart from ANNs, have used formalisms for representing moral rules that reduce the ability of the system to reason flexibly. Since natural language is used to express moral arguments and reasoning, it may be better to represent moral principles and reasoning in ways that are more naturalistic than formalistic. Fuzzy logic is a representation method for modelling logical reasoning based on fuzzy sets [47] and is particularly useful for handling natural language reasoning [48].

Consider the following scenario. An IAV is currently under the control of a human driver, who is driving along a multi-lane highway at 50 mph. Three sensors provide data on distance to the vehicle in front, keeping within the lane and current speed. The data from these three sensors are monitored constantly. The *sensor technology* question is under what physical circumstances (distance to next vehicle, location within lane, speed) the IAV should take over control if there is danger of an accident in order to prevent or reduce the impact of the accident. The *ethical* question is when is it morally acceptable to take over control if there is danger of an accident. The moral question may need to take into account respect for the driver's autonomy and consequences of taking over in a way that the sensor technology question does not. A related ethical question is, after taking over control and implementing remedial action, under what circumstances is it morally justified to hand back control to the human driver so that the autonomy of the driver is respected, even if the driver does not take back control.

To address these ethical questions, we start with two very general and non-controversial virtuous moral principles (VMPs) for an IAV:

*VMP1: If it is right to take control and if it is good take control, then the virtuous outcome is to take control.*

*VMP2: If it is wrong to take control and if it is bad to take control, then the virtuous outcome is to not take control.*

In this scenario, right and wrong are related to the duty to take control in a dynamic driving situation as given by sensor readings, and good and bad are related to the effects of taking control on the autonomy of the human driver in that situation. So we hypothesize two reasoners: a 'right/wrong' reasoner for dealing with the right/wrong or duty dimension, and a 'good/bad' reasoner for dealing with the 'good/bad' or utilitarian dimension. The 'right/wrong' and 'good/bad' reasoners reflect very generally deontological and consequentialist approaches, respectively. VMP1 and VMP2 are virtue ethics principles for an ethical IAV that has as its main function the taking into account of both the deontological and consequentialist aspects of the situation through rational decision making.

## 4. Methods
### 4.1 Representations

Right and wrong actions, as well as good and bad actions, need to be described in terms of risk. For the right/wrong reasoner ('right_wrong'), the three sensors are represented through the fuzzy variables of *distance*, *lane* and *speed*. Each has high-risk and low-risk values that can be described through membership functions (MFs), where the horizontal axis specifies all values of the set and the vertical axis the risk value from 0 (not at all) to 1 (totally). For instance, for *speed*, the sensor readings can go from 0 mph to 100 mph on the horizontal axis, and the two risk values of 'low risk' and 'high risk' are as shown in Figure 1. Low risk (in red) is identified as having maximal membership value 1 until 40 mph, at which point its membership value drops steadily to 0 at 80 mph. High risk (in black) has 0 membership value until 40 mph, at which point its membership value increases steadily to 80 mph, at which point it has maximum membership value. These trapezoidal-shaped membership functions ('trapmf' under Current MF Type in Figure 1) can be differently set for different variables. For instance, for *lane*, low risk (Figure 2, in red) can retain maximal membership value until sensor reading 8, at which point it drops sharply to minimal membership value at sensor reading 9 (Figure 2). High risk, (in black) on the other hand, is minimal until reading 7, at which point it rises sharply to maximal at reading 8. The *lane* variable is assumed to return high risk readings for proximity to edges of the lane and low risk readings for being central in the lane.

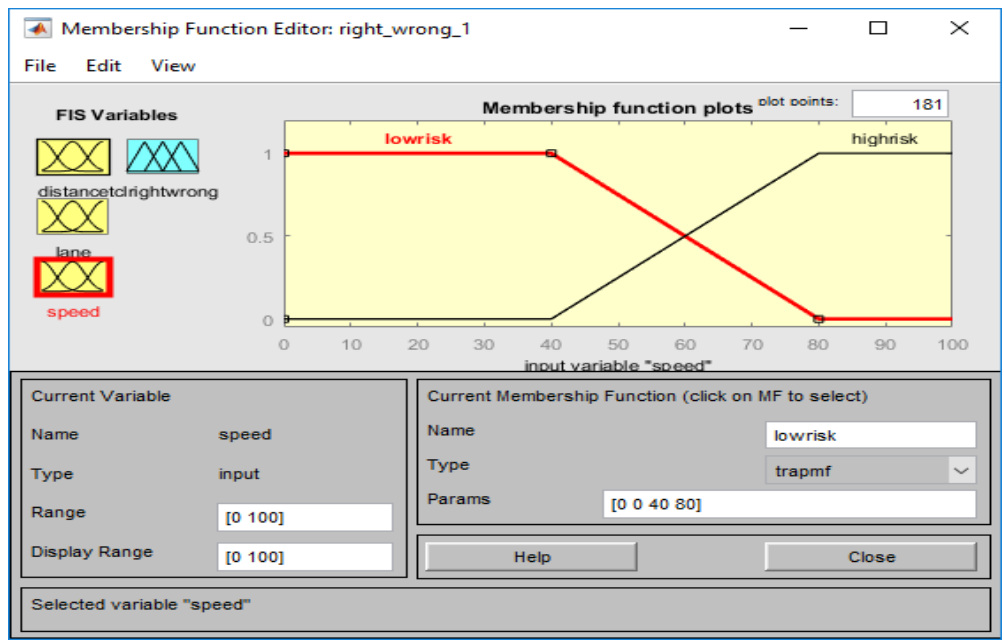

*Figure 1: Fuzzy membership functions (MFs) for low risk and high risk for the variable 'speed' as part of the right/wrong reasoner,, as depicted using MATLAB's fuzzy logic toolbox. At the upper left of the figure are the three input variables 'distance', 'lane; and 'speed' (with 'speed' highlighted to show its MFs) corresponding to three sensors, as well as the output variable 'tcrightwrong' (for 'take control right or wrong'). The horizontal axis for 'speed' consists of mph readings ('range') and the vertical axis describes the amount of set membership for the two MFs of 'lowrisk' and 'highrisk', both of which have trapezoidal shape ('trapmf'). The parameters box describes the initial, second, third and fourth points of the trapezoid (in this case, for 'low risk' highlighted in red).*

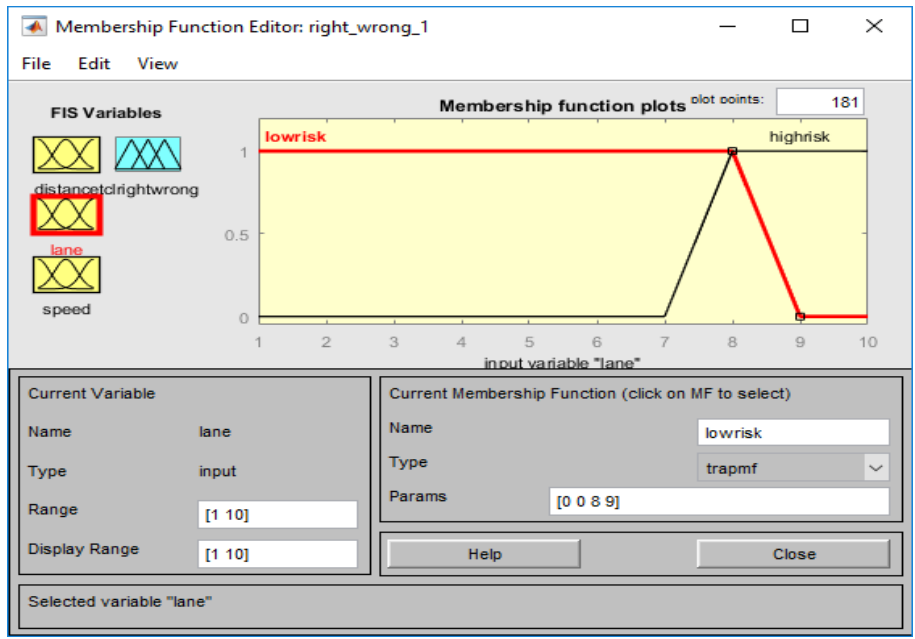

*Figure 2: Fuzzy MFs for 'lane' of the right/wrong reasoner, indicating low risk being minimal until sensor reading 8 and dropping to minimal at sensor reading 9, whereas high risk is minimal until sensor reading 7, at which point it rises sharply to maximal at sensor reading 8.*

The Supplementary Information (SI) appendix contains the MFs for the third variable *distance* for the right/wrong reasoner (Table 1, SI), where low risk is maximal until value 5 and minimal for value 6, and high risk is minimal until 5 and maximal from level 6. The right/wrong reasoner also contains an output variable labelled *tcrightwrong* (for 'take control right or wrong'), which has the two MFs: one for 'take control is right' ('tcright') and the other for 'take control is wrong' ('tcwrong'). These MFs can be specified as sigmoidal rather than trapezoidal (Figure 3), where 'tcwrong' is maximal until value 7 and drops gradually to minimal at value 10 (falling sigmoid MF, or zmf), and 'tcright' is minimal until value 4 and then rises gradually to value 7, where it is maximal (rising sigmoid MF, or smf). The outcome of this right/wrong reasoner is specified to be in the range of 1 to 10). These MFs can be changed to other values and shapes if necessary.

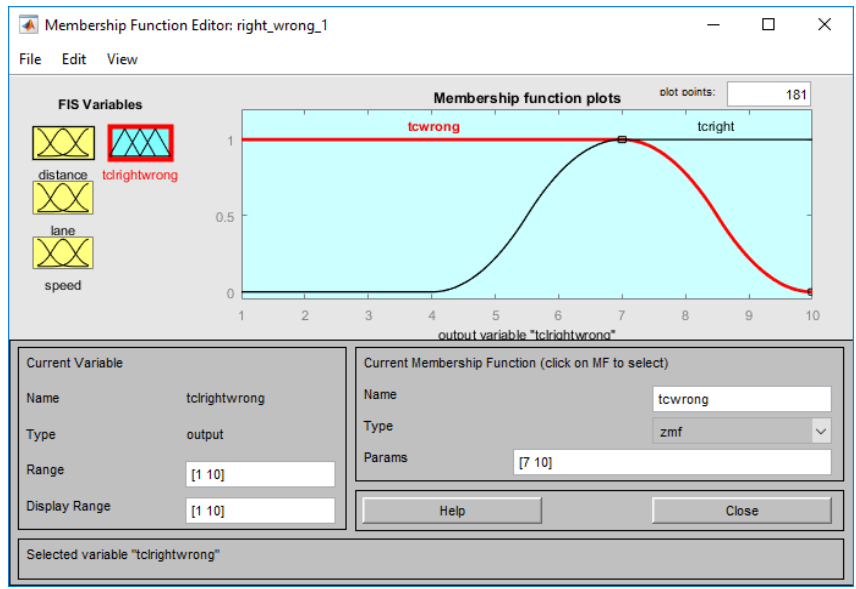

*Figure 3: MFs for the moral output of right and wrong on a scale of 1-10, with 'take control is wrong' (tcwrong) being maximal until value 7 (in red) and 'take control is right' (tcright) being minimal until 4.5 (in black).*

### 4.2 Architecture

As noted above, two ethical reasoners are used at the first level of the ethical reasoning architecture: one for (deontological) right/wrong, and the other for (consequentialist) good/bad. Both reasoners use the same three input variables *(distance*, *lane*, *speed*) and MFs (high risk, low risk), but the use of the same variables and MFs in each reasoner need not necessarily be the case. For our example, the MF parameter values for the good/bad reasoner are different from those for the right/wrong reasoner (Supplementary Information (SI), Table 1).

Each reasoner has general principles for producing moral output from their inputs. For the right/wrong reasoner, two right/wrong principles (RWPs) are:

*RWP1. If distance is low risk, lane is low risk and speed is low risk then take control is wrong.*

*RWP2. If distance is high risk, lane is high risk and speed is high risk then take control is right.*

These two principles are minimal and morally non-controversial. They describe the extreme circumstances under which taking over control from the human driver is right or wrong.

Similarly, two minimal and non-controversial principles are used for the good/bad reasoner (GBPs), with the outcomes being taking control is bad and taking control is good:

*GBP1. If distance is low risk, lane is low risk and speed is low risk then take control is bad.*

*GBP2. If distance is high risk, lane is high risk and speed is high risk then take control is good.*

These consequentialist principles take into account the effect on the driver's autonomy. The standard fuzzy conjunction operator of minimum value is used to calculate the value of the consequent and hence the outcome for all principles.

### 4.3 Forming an ethical judgement

The virtuous meta-ethical controller (VMEC) at the next level takes as input the two moral outcomes of right/wrong and good/bad to produce a summative moral judgement. The two principles of the VMEC are VMP1 and VMP2 (as described earlier):

*VMP1: If it is right to take control and it is good take control, then the virtuous action is to take control.*

*VMP2: If it is wrong to take control and it is bad to take control, then the virtuous action is to not take control.*

There are two VMEC membership functions for each input specified not in terms of risk but in terms of rational outcome. For right/wrong, the two membership functions ('right wrong don't take control', rwdtc; 'right wrong take control', rwtc) are specified as decreasing and increasing sigmoid functions (Figure 4), whereas for good/bad the two membership functions ('good bad don't take control, gbdtc; 'good bad take control', gbtc) are specified as trapezoidal for the purpose of example (SI, Table 1). The output is 'virtuous control decision', which has two membership functions: virtuous not to take control ('vcno') and virtuous to take control ('vcyes'). These are specified as falling and rising sigmoids around the mid point 5.5 on a scale of 1-10 (Figure 5).

The two principles for the VMEC (VMPs) are virtuous because they make the VMEC a good controller by taking deontological and consequentialist aspects into account when deciding on the best course of action so that controller best fulfils its function.

The overall architecture is given in Figure 6.

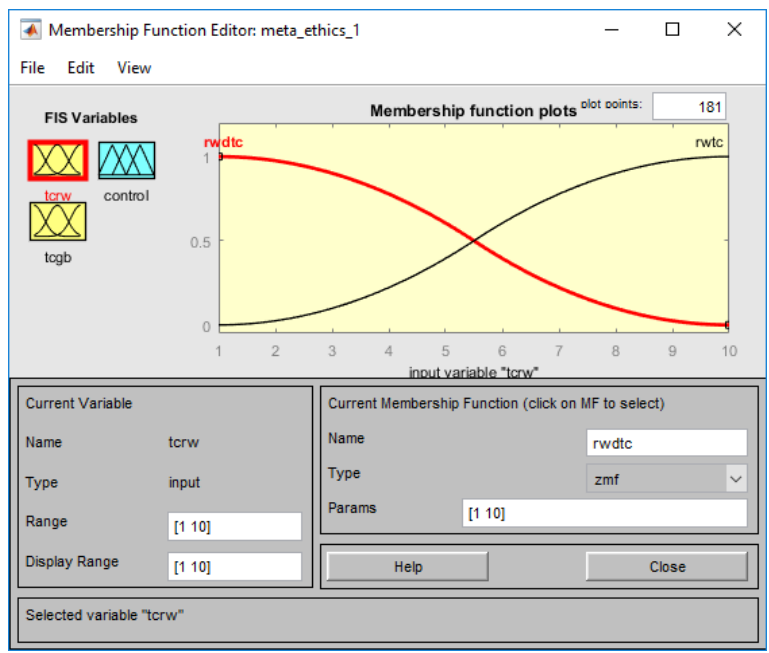

*Figure 4: First input to the virtuous meta ethics controller (VMEC) consisting of the output from the right/wrong dimension, consisting of two sigmoidal membership functions 'right wrong do not take control' (rwdtc) and 'right wrong take control' (rwtc)*

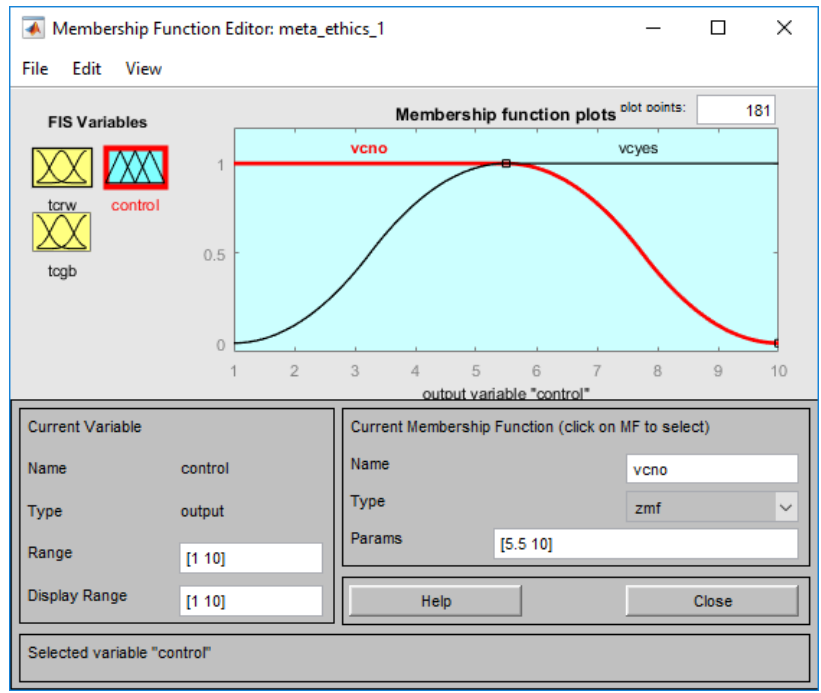

*Figure 5: The output from the VMEC is specified through two sigmoidal membership functions "virtuous not to take control" (vcno, in red) and "virtuous to take control" (vcyes, in black)*

### 4.4 Producing a crisp output

A key aspect of all fuzzy systems is the method used to produce 'crisp' output based on membership functions. The most common output calculation method involves centroids (calculating the centre of gravity of the summed combined subareas of membership functions) using the Mamdani method of aggregated output distribution [49], which is the method used here.

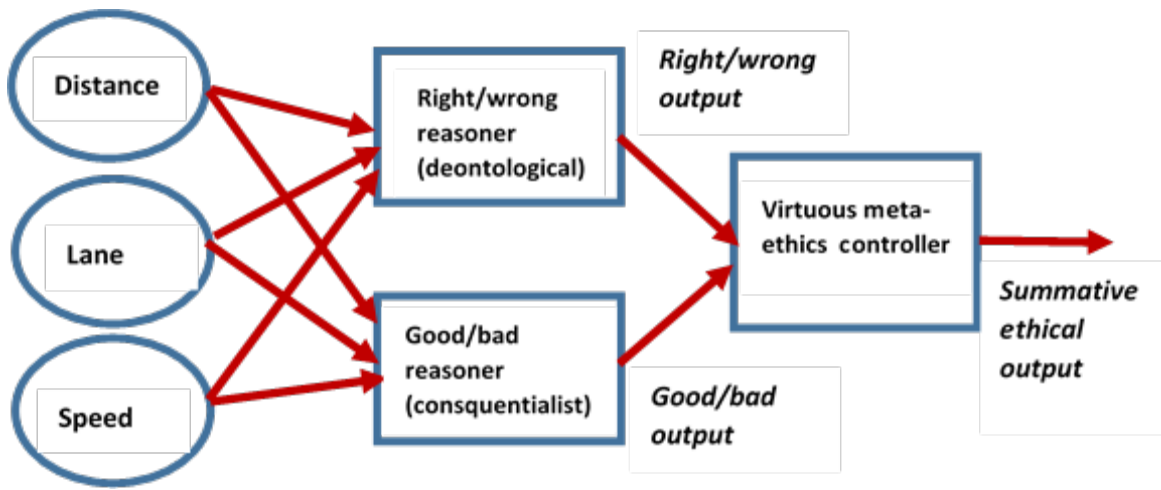

*Figure 6: Overall architecture of fuzzy ethical reasoning system, where sensor data concerning distance, lane adherence and speed are input to the right/wrong (deontological) reasoner and the good/bad (consequentialist) reasoner, each of which produces output values for the virtuous meta-ethics controller to produce summative ethical output values. Each reasoner and controller contains two ethically uncontroversial principles (six principles in total).*

All the parameters used so far are chosen for demonstration purposes, and it is up to designers of IAV ethical system designers to use the most appropriate variables, the number and type of membership functions as well as the principles depending on the environment in which the IAVs are intended to be used. The key point is that all moral principles are minimal, uncontroversial and sensible for any IAV to adhere to. The use of fuzzy logic ensures that there is no over-reliance on programming rules for every particular situation.

After the model is designed and developed, simulations across a range of input values can determine the effectiveness of the design choices as well as produce moral output that can be used to learn ethical rules without human direction, as will be seen below.

MATLAB R2015b and Simulink were used for all simulations, and SPSS v23 (cluster analysis, regression) and Weka 3.6.13 (ANN, rule induction) for data analysis and machine learning.

## 5. Experiments and results
**5.1 Ethics of taking control**

The behaviour of this ethical system was simulated continuously for 10 time units by setting the speed sensor to go through one full cycle of speeds from 0 to 100 mph, the lane sensor to go through two full cycles from 1 to 10 and the distance sensor to go through four full cycles from 1 to 10. At the end of the simulation, three streams of outputs are produced for each ethical reasoner based on their separate membership functions (Figure 7). Step-changes in the VMEC output indicate possible boundaries of steady states (output values do not change for a certain period of time), attractor points (convergence of states) and periodic points (repeating states).

Two step-changes about the 5.5 value on the y-axis (Figure 7, white line) indicate three possible VMEC steady state outcomes in the system. The VMEC outputs a value greater than 6 during the early stages of the simulation when all three sensors indicate high risk. This can be interpreted as the IAV taking over control being the virtuous outcome (class 2 ethical outcome). Where the VMEC returns values above 5 but below 6, there appear to be mitigating circumstances (e.g. low risk of collision, low risk of lane straying and/or low speed) that indicate a grey state (class 1 ethical outcome). For VMEC values below 5, the outcome is that there is no virtuous ethical reason for taking over control (class 0 ethical outcome).

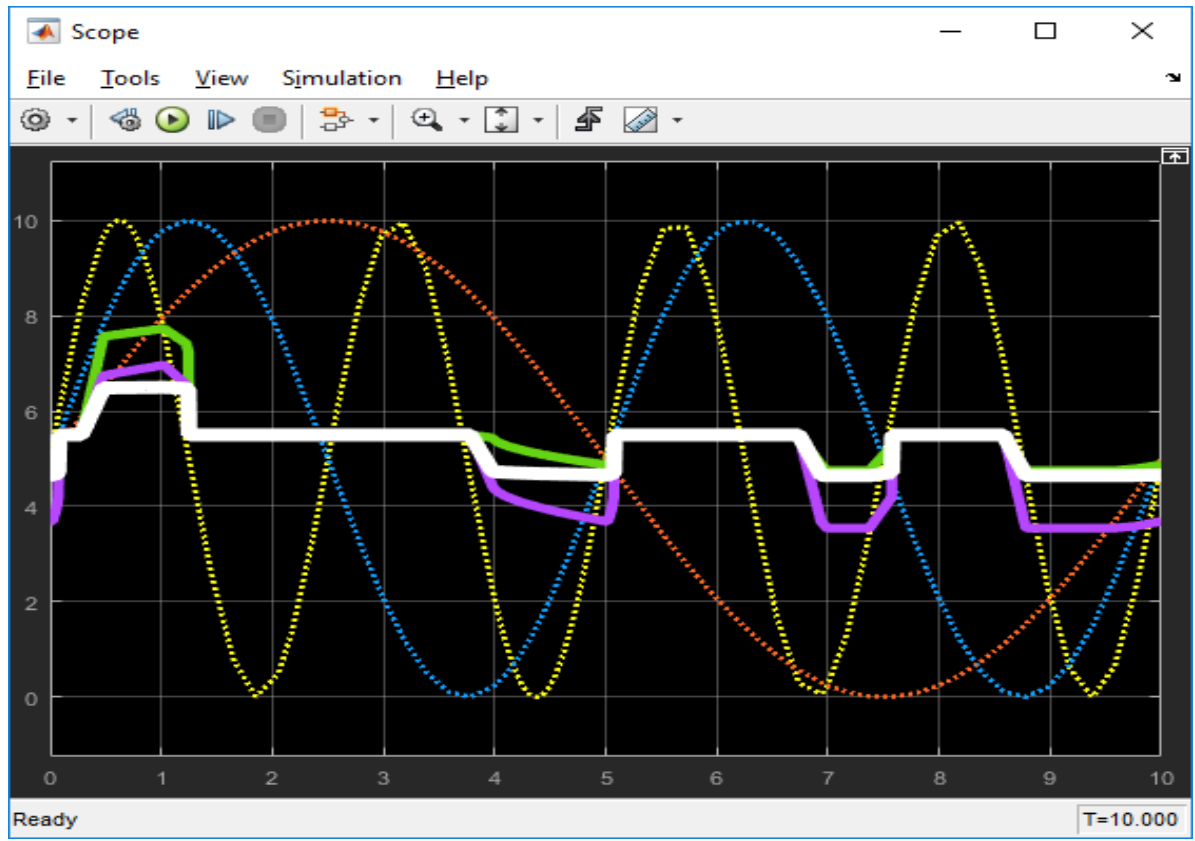

*Figure 7: Streamed output of virtual meta-ethics controller (VMEC, thick white line) for the three input sensors of speed (orange dotted line, one full cycle across 10 time units), lane adherence (blue, two cycles) and distance (yellow, four cycles), with the corresponding streamed output of the right/wrong reasoner in thick green and the good/bad reasoner in thick purple. The y axis represents both the range of input values and output fuzzy values, but note that the speed input values (varying between 0 and 100 mph) have been multiplied by 0.1 to fit them on the graph for the sake of legibility.*

Variable time-step sizing over 10 time steps, where the step size is reduced when the model changes rapidly and increased when the model is static, resulted in 227 of the 308 samples (73.7%) being in class 0 (VMEC output values below 5), 20 (6.5%) in class 1 (between 5 and 5.99) and 61 (19.8%) in class 2 (6 and above).

Hierarchical cluster analysis (HCA) of the six continuous variables using squared Euclidean distance as the similarity measure and centroid clustering as the method for separating clusters showed that the output of the VMEC was closer to (more associated with) the right/wrong output (distance of 104.47) than the good/bad output (137.93). The average VMEC steady state class values were 4.69 for class 0, 5.50 for class 1 and 6.48 for class 2.

NNge (nearest neighbour with generalization) is a rule induction algorithm [28] that merges exemplars and forms hyperreactangles in feature space [29] to produce rules containing ranges of values for continuous variables. NNge is known for producing verbose and extensive rule sets for maximum knowledge extraction. NNge using the five continuous variables (three sensor and two ethical reasoner streamed data) correctly classified all 308 samples into the three discrete virtuous classes using the following learned virtuous moral rules (VMRs):

VMR1: IF  (0.0 ≤ distance ≤ 6.0)  AND (0.0 ≤lane ≤ 8.25) AND (0.0 ≤ speed ≤ 79.9)  AND (4.75 ≤ goodbad_output ≤ 5.46)  AND (3.54 ≤ rightwrong_output ≤ 4.47) THEN virtuous ethical class 0.

VMR2: IF  (0.48 ≤ distance ≤ 9.86) AND (goodbad_output = 5.5) AND (rightwrong_output = 5.5) THEN virtuous ethical class 1 (grey state).

*VMR3: IF (5.0 ≤ distance ≤ 10.0) AND (8.04 ≤ lane ≤ 10.0) AND (66.1 ≤ speed ≤ 85.4) AND (7.03 ≤ goodbad_output ≤ 7.73) AND (6.04 ≤ rightwrong_output ≤ 6.97) THEN virtuous ethical class 2.*

VMR3 provides the metric-based ethical rule that taking over control from the human is virtuous if the distance risk is between 5 and 10, the lane risk is between 8 and 10 and the speed is over 66 mph. In this case, the good/bad dimension is over 7 and the right/wrong dimension over 6. The upper bounds can be ignored since they reflect the maximum values in the sampled data. VMR1 states that there is no virtuous ethical reason to take over if the good/bad dimension is below 5.46 and the right/wrong dimension is below 4.47. The lower bounds can be ignored since they reflect the minimum values in the sampled data. VMR2 is the default grey state between the two other states and is specified by the deontological and consequentialist outputs both being neither too high nor too low (= 5.5). Ten-fold cross-validation holding back a random 10% of samples in each fold also produced 100% correct classification, indicating that the learned rules are effective and reliable under NNge.

A perceptron with the five variables (three sensor, two ethical) as input, three hidden units and three output nodes (one for each class of VMEC outcomes), with learning rate 0.3, momentum 0.2 and 500 training epochs, successfully learned all 227 class 0 cases and 61 class 2 cases but wrongly classified all class 1 samples as class 2. A series of ANN experiments using a perceptron (three sensors and two level 1 ethical reasoners as input, three VMEC classes as output) showed that the 'minimum' architecture required for successful learning of all 308 instances required a learning rate of 0.1, a momentum of 0.2, 1000 training epochs and 4 hidden nodes. 10-fold cross-validation using these parameters resulted in an overall 95.13% accuracy, with all class 0 and class 2 samples correctly classified but only 5 of the class 1 samples correctly classified (6 classified as class 0 and 9 as class 2). Increasing the training epochs to 10,000 resulted in further improvement to 98.43% accuracy (2 class 1 wrongly classified as class 2). Adding additional hidden units did not improve the performance. In other words, the ANN had difficulty in generalizing to the grey class, possibly due to the its relative under-occurrence in comparison to the other two classes.

Linear regression with the continuous VMEC output values as the dependent variable and the five other variables (three sensor, two ethics) as independent variables resulted in an $R^2$ of 0.97 (i.e. 97% of variance in the dependent variable explained) and the following model:

*2.208 (constant) + 0.67 (goodbad) + 0.17 (lane) + 0.22 (rightwrong) + 0.059 (speed) −0.09 (distance)*

The highest coefficient (0.67) is for goodbad, indicating that, according to regression, this has the most influence.

While VMR1-VMR3 focus on identifying circumstances under which it is ethical for an IAV to take over control, the rules can also be adapted to identify circumstances under which the IAV should, morally, hand control back to the human. Each of VMR1-VMR3 can be amended so that the conclusions are interpreted as staying in the same ethical state or moving to a new ethical state if in a different state. So VMR1, for instance, can be interpreted as being ethical to hand back control to the human if the immediately previous moral state was not class 0.

**5.2 Ethical dilemmas**

Many aspects of the architecture can be used for adapting the moral reasoner to deal with dilemmas. As noted earlier, moral dilemmas are characterized by morally right (deontological) action leading to morally bad (utilitarian or consequentialist) outcomes, and vice versa. With dilemmas, rightness and wrongness must be balanced against goodness and badness to reach justified judgements.

Consider the situation where an IAV is confronted by a child crossing the road in front of it and it is too late to brake without hitting the child. It must decide whether to brake and take avoidance action by swerving onto the pavement where almost certainly, according to its sensors, an adult will be killed, or whether to keep going straight. In this situation, there is no virtuous outcome where someone is not going to be badly injured or die. One moral decision is to justify swerving onto the pavement on the basis that the life of a child counts more than the life of an adult.

To address this dilemma, the IAV needs two sensors to provide information concerning risk of death by driving straight ahead and risk of death by taking swerving action. These two sensors feed into the right/wrong reasoner that has two outputs for swerving being wrong and serving being right. A third sensor identifies the age of any pedestrian straight ahead. SI, Table 2 provides details of the membership functions and Figure 8 an overview of the ethical dilemma architecture. Figures 9, 10 and 11 provide visualizations of all the variables used in the dilemma architecture.

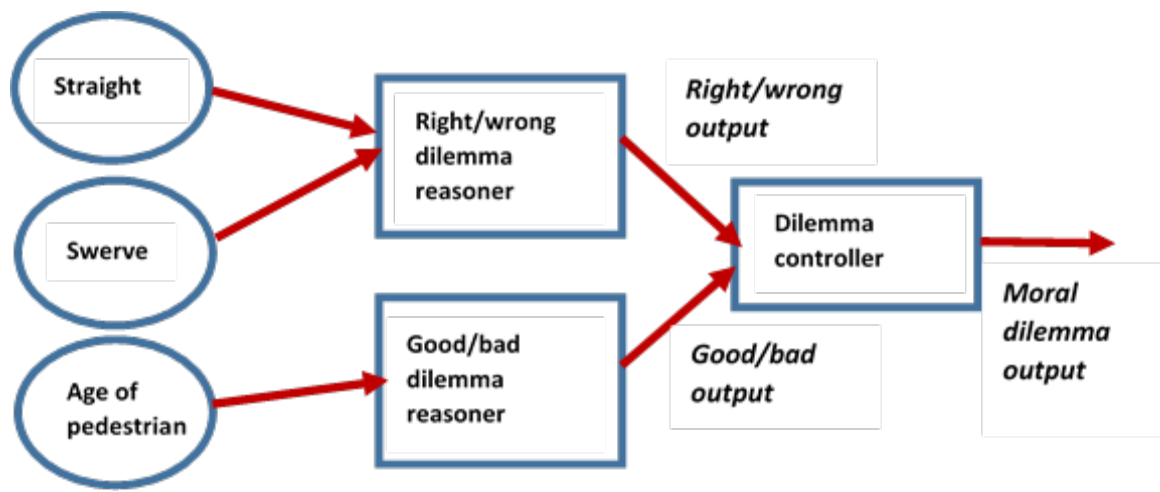

*Figure 8: Moral dilemma architecture, with sensors for identifying what is straight ahead and what is on the pavement for the right/wrong reasoner, and a sensor for pedestrian age for the good/bad reasoner, with the dilemma controller providing various dilemma outputs for varying sensor values.*

The principles for each of the dilemma reasoners are uncontroversial. For the right/wrong reasoner, the two dilemma principles (RWDPs) are:

*RWDP1: If going straight ahead leads to high risk of death and swerving leads to low risk of death, then swerving is right.*

*RWDP2: If going straight ahead leads to low risk of death and swerving leads to high risk of death, then swerving is wrong.*

For the good/bad dilemma reasoner, the principles (GBDPs) are:

*GBDP1: If pedestrian is young then taking avoiding action is good.*

*GBDP2: If pedestrian is not young then not taking avoiding action is not bad.*

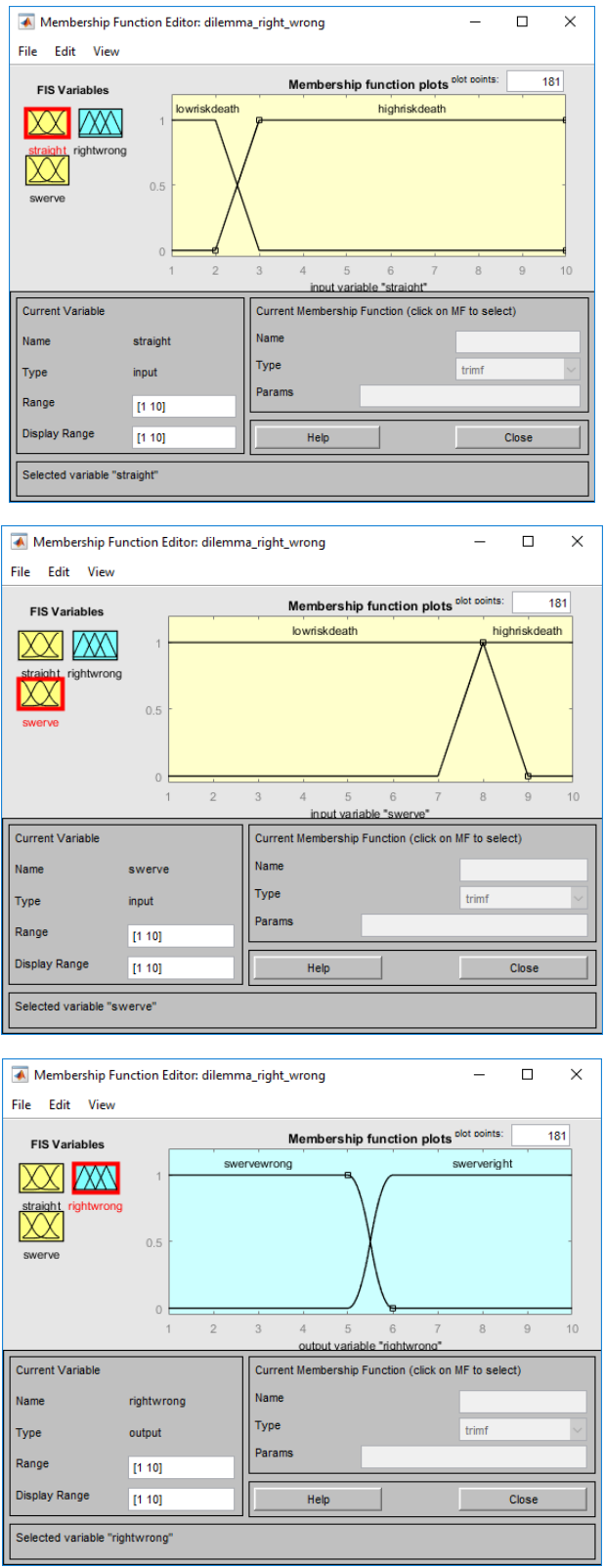

*Figure 9: The two input variables of the right/wrong dilemma reasoner, with straight ahead (top) and swerve (middle) each containing two MFs 'lowriskdeath' and highriskdeath', and output variable (bottom) with two MFs 'swervewrong' and 'swerveright'. All MF parameters are specified in S1, Table 3.*

GPDP2 is not a licence to run over older pedestrians but is a principle to be interpreted in the context of no other action being possible when faced with pedestrians who will be injured no matter what course of action one takes.

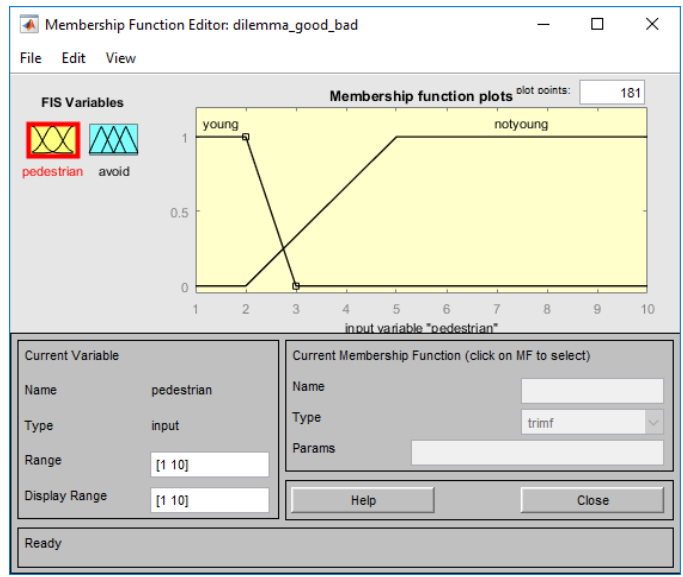

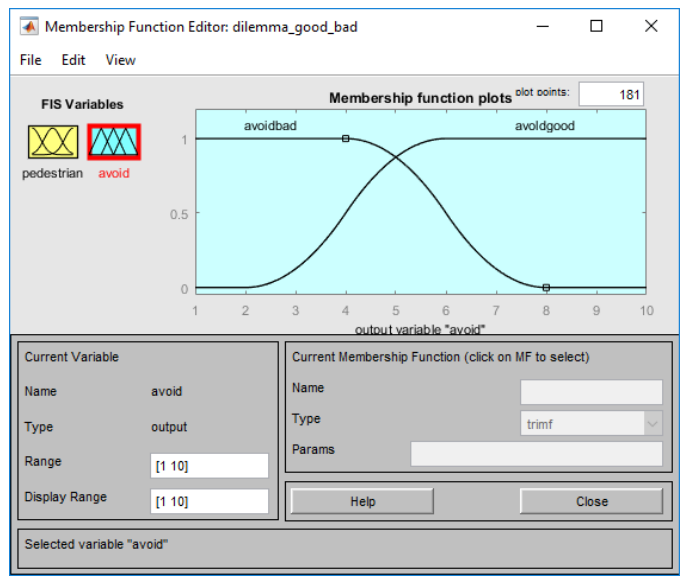

*Figure 9: The input and output variables of the good/bad dilemma reasoner, with 'pedestrian' (input, top) having two MFs (young, old) and 'avoid' (output, bottom) have two MFs (avoidgood, avoidbad). All MF parameters are specified in S1, Table 2.*

The ethical dilemma controller (EDC) takes as inputs the two outputs from the right/wrong and good/bad reasoners and combines them into an output using the following two ethical dilemma principles (EDCPs):

*EDCP1: If deathrisk is high and pdestrianrisk avoid is good then brake but swerve.*

*EDCP2: If deathrisk is low and pedestrianrisk avoid is bad then brake and continue straight ahead.*

For simulation purposes, the signal generators for straight, swerving and pedestrian age (multiplied by 0.1 to fit the values on the graph) vary from 1 to 10, and with one full cycle, two full cycles and four full cycles, respectively. The simulation is run this time with fixed steps in discrete time to ensure that the greatest variety of inputs to outputs (Figure 11).

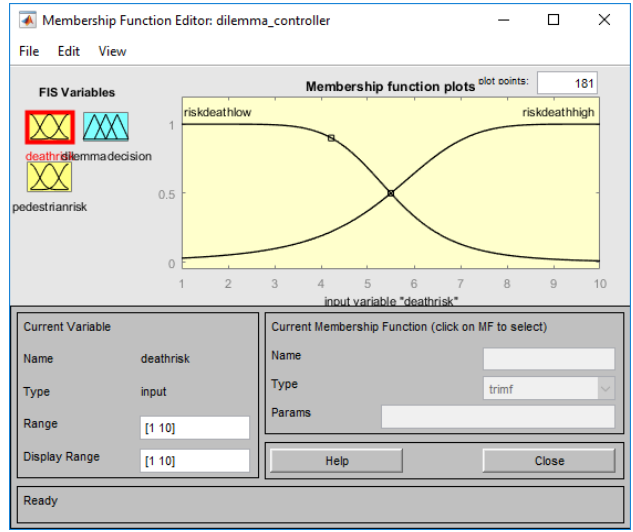

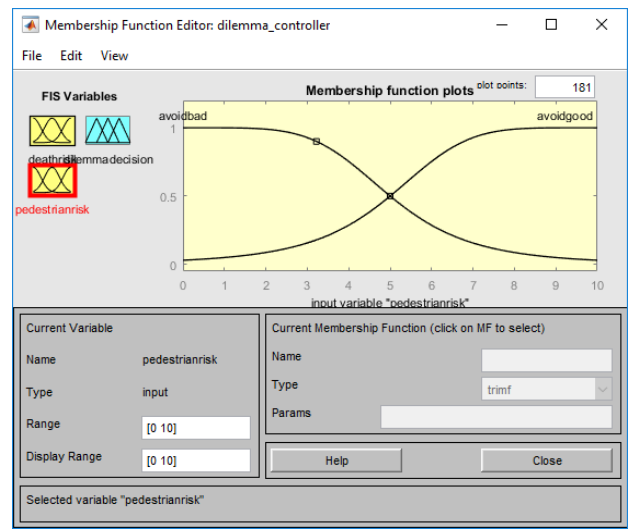

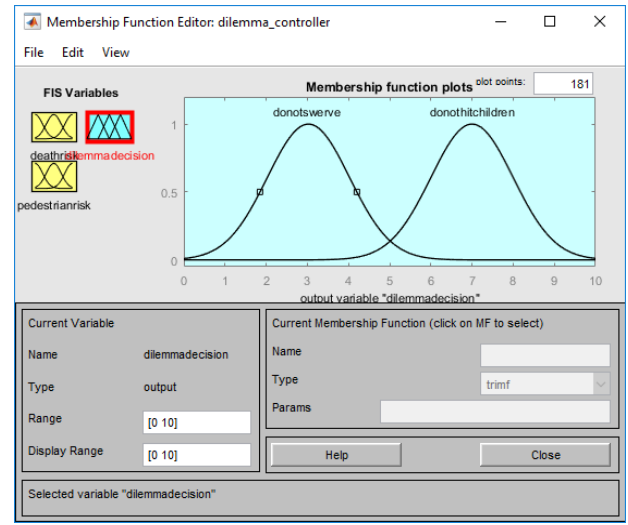

*Figure 10: The two input variables and one output variable of the dilemma controller, with deathrisk (input, top) having two MFs (riskdeathlow, riskdeathhigh), pedestrianrisk (input, middle) two MFs (avoidbad, avoidgood) and dilemmadecision (output, bottom) two MFs (donotswerve, donothitchildren). All MF parameters are specified in S1, Table 2.*

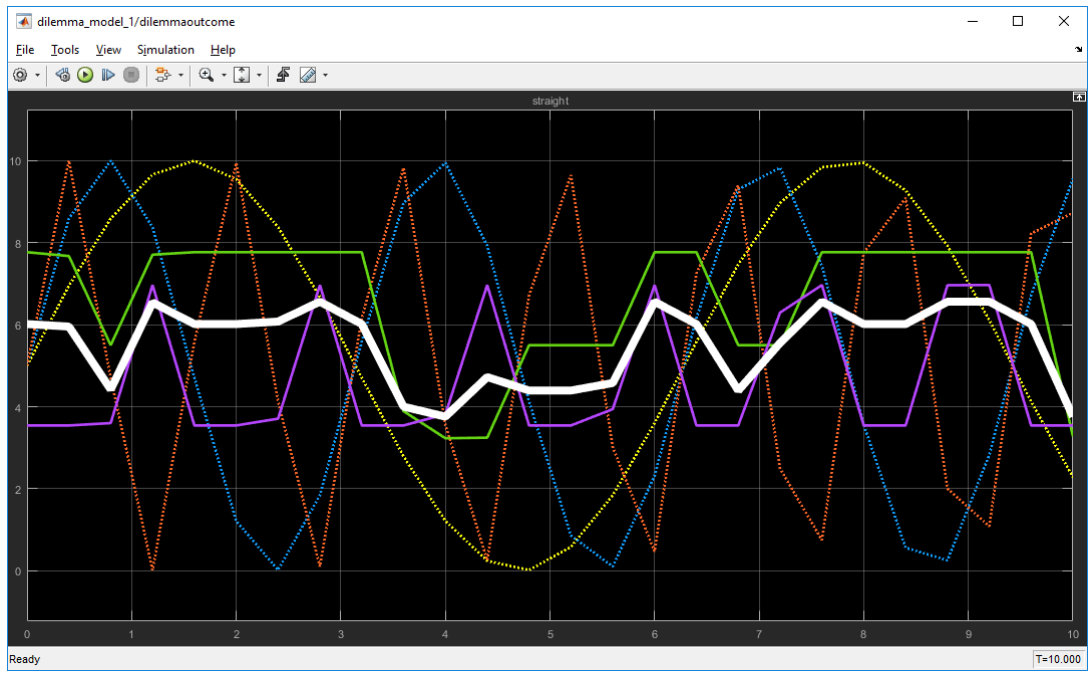

*Figure 11: Output of dilemma controller (thick white line) for the three input sensors of straight direction (yellow dotted line), swerve (blue dotted line) and pedestrian age (orange dotted line), with the corresponding outputs of the right/wrong reasoner in green and the good/bad reasoner in purple. The y axis represents both the range of input values and output fuzzy values, but note that the age input values (varying between 0 and 100) have been multiplied by 0.1 to fit them on the graph for the sake of legibility.*

26 samples were produced, with the dilemma output having a mean value of 5.52 and median of 6.02. Using the median value as the cut point to distinguish class 1 ('brake and go straight ahead') and class 2 ('brake and swerve'), NNge produced the following ethical rules (EDRs) for addressing the dilemma:

*EDR1: IF (30.1 ≤ pedestrian age) AND (3.23 ≤ swerve_rightwrong ≤ 7.77) AND (3.54 ≤ avoid goodbad ≤ 3.94) THEN the appropriate moral outcome is brake and go straight ahead*

*EDR2: IF (3.6 ≤ risk of death by going straight ahead) AND (0.2 ≤ pedestrian age ≤ 20) AND (7.71 ≤ swerve rightwrong) AND 6.96 ≤ avoid goodbad)  THEN the appropriate moral outcome is brake and swerve.*

*ER3: IF (7.92 ≤ risk of death by swerving) AND (2.6 ≤ pedestrian age ≤ 24.9) AND (3.25 ≤ swerve rightwrong ≤ 5.5) AND (6.3 ≤ avoid goodbad) THEN the appropriate moral outcome is brake and go straight ahead.*

EDR1 states that if the pedestrian in front is over 30, if swerving has moderate to high risk of death (between 3.2 and 7.8) and there is not enough good reason to take avoiding action (between 3.5 and 4), then the morally correct action is to go straight ahead in this dilemma.

EDR2 states that the morally correct decision when confronted by a risk of death of greater than 3.6 by going straight ahead and when the pedestrian in front is below 20 years of age is to swerve, given that swerving is good in this situation (i.e. above 6.3) even if it is not totally right to swerve (swerving is right value of between 3.5 to 5.5).

However, EDR3 modifies EDR2 by adding that if the risk of death is very high for swerving and the pedestrian is between 2.6 and 25 years old, but swerving is right is not greater than 5.5 even if it is good to swerve (above 6.3), then the morally correct action in this dilemma is to brake and go straight ahead.

While some may argue that it may always be better to swerve to hit a child, an IAV that has the dilemma reasoner above encapsulates this moral dictum to some extent but not to every extent.

## 6. Discussion

The aim of the research was to identify whether a fuzzy ethical reasoner could generate its own data for learning ethical rules to help it determine when it was ethical to take over control or hand control back to humans. The results above indicate that it is possible to introduce minimal and non-controversial moral principles in three different moral dimensions: the right/wrong dimension, the good/bad dimension, and the meta-ethics virtue dimension of combining the two previous moral dimensions. The system is 'autometric' because, through simulation using fuzzy logic, the ethical reasoner can generate its own data for learning moral rules and models without interference or bias from humans. Once an initial set of rules or models is derived, the ethical reasoner can use these as a benchmark for refining or experimenting with variations in the same or other risk membership functions to help it determine other possible circumstances under which it is ethical to take over control or hand it back. For instance, HCA showed that the VMEC output was closer to the deontological dimension (right/wrong) than the consequentialist (good/bad) dimension. While this may be appropriate for IAVs, IAMs in other situations (e.g. robots helping to nurse aged people) may have membership function shapes and values that lead to stressing the utilitarian aspects of respecting autonomy more than the deontological so that humans do not feel controlled.

The architecture of Figure 6 can be totally separate from the actual control architecture of an IAM. The actual control architecture of an IAV, for instance, is likely to contain many more sensors and much faster sampling rates than those required by an ethics architecture. Three possibilities exist. The first is for the ethics architecture to be totally independent of the control architecture (zero ethics coupling). The architecture provides an ethical commentary to the sensor-based behaviour of the IAM but does not interfere in any way with the actual behaviour of the IAM. Such commentary could be used off-line to monitor and evaluate AIM actions in responses to sensor data. The second is for the ethics architecture to work in parallel with the control architecture of an IAM so that control architecture output and ethics architecture output are combined in some way (parallel ethics coupling) for 'morally considered' action. And the third possibility is for the output of the control architecture to be one of the inputs to the ethics architecture (serial ethics coupling) so that no action is possible without 'ethical approval'. The choice of architectural involvement will depend on the context in which IAMs will be used. Perhaps sometime in the future there will be a requirement for all IAMs that have the potential to inflict harm on humans to have a compulsory ethics architecture so that control information and ethics outcomes can be used together in parallel or in serial mode.

When dealing with ethical dilemmas that trade off aspects of deontology with consequentialism, the results again indicate that adopting a fuzzy logic approach can produce a spectrum of outputs that can be used to learn what to do when confronted by such dilemmas. Not everyone will agree with the decisions made by an IAV in these situations, but that is in the nature of dilemmas. The important point here is that the IAV can provide reasons for its decisions, which can in turn lead to refinement through improved membership functions and the introduction of more fuzzy variables if appropriate. Transparency of reasoning is critical in resolving moral dilemmas, even if the outcomes are not universally agreed.

We have also shown how the ethical reasoner, through a fuzzy logic approach, achieves interactivity, autonomy and adaptability. Interactivity is achieved through dynamic processing of sensor information to ensure that moral judgements are continuously made and monitored. Autonomy is achieved through the system internally processing data from sensors and simulations using principles but deriving rules from this internal processing without human intervention. Adaptability is achieved through simulations of dilemmas that allow the system to trade off aspects of right/wrong against good/bad.

None of this is achievable without human designers providing the first set of principles and fuzzy variables using a variety of membership functions. But human designers do not have to specify every possible ethical situation using the approach described here and can leave the derivation of appropriate rules to the ethics architecture. As noted by several researchers, just because fuzzy logic deals with inexactness and approximation due its use of the real number interval between 0 and 1, that does not mean that fuzzy logic cannot be subject to formalization of its inferential processes [52] so that fuzzy reasoning is shown to be effective and computable [53]. By limiting the involvement of human designers to the input of uncontroversial principles and an initial choice (and shape) of membership functions to be used for representing those principles, the danger of introducing biases and prejudices is minimized. The only human input initially is in the first set of principles and risk membership functions to help the autometric ethical IAM establish a base from which to learn enhanced models of ethical reasoning. There is no reason why an IAM should not choose different membership functions and shapes to experiment with through evolutionary algorithms, for instance.

Finally, research into how values can be embedded into autonomous intelligent systems through advances in data collection, sensor technology, pattern recognition and machine learning is now actively encouraged as a method of achieving a correct level of trust between humans and autonomous intelligent systems [54]. Ethical reasoning in philosophy and machine ethics has so far dealt almost exclusively with qualitative reasoning. The use of intervals and values in the range of 0 and 1 in ethical output is not familiar to us humans and could lead to accusations of a metric-based ethics being 'ethics by numbers'. But we have shown that it is possible to assign categories to the output of autometric reasoning for class-based or categorical learning if required. But more importantly, if we want our IAMs to develop a sense of right and wrong so that they do not harm us, we may have to accept that non-qualitative, metric-based ethics is the best way to go.

### Model code and data availability

The MATLAB code for all fuzzy logic reasoners and the Simulink simulation, as well as the data produced by the simulation for use in data mining and statistical analysis, will be made available by emailing the corresponding author from an academic email address.

# References

1. Long, LN, Hanford SD, Janrathitikarn O, Sinsley GL, Miller JA. A review of intelligent systems software for autonomous vehicles. *Proceedings of the 2007 IEEE Symposium of Computational Intelligence in  Security and Defense Applications* (CISDA 2007), 2007, 69-76.

2. Brooks RA. A robust layered control system for mobile robot. *IEEE Journal of Robotics and Automation. 1986,* 2(1): 14-23.

3. Brooks RA. Intelligence without representation.  *Artificial Intelligence Journal*. 1991, 47: 139-159.

4. Brooks RA. A robot that walks; Emergent behaviour from a carefully evolved network. *Neural Computation*. 1989, 1 (2): 253-262.

5. Matijevic M. Autonomous navigation and the Sojourner Microrover. *Science.* 1998, 280(5362): 454-455.

6. Breazeal C. *Designing Sociable Robots.* 2002, MIT Press.

7. Arkin RC and Balch T. AuRA: Principles and practice in review. *Journal of Experimental and Theoretical Artificial Intelligence*. 1997, 9(2-3): 175-189.

8. Stover JA, Ratnesh K. A behaviour-based architecture for the design of intelligent controllers for autonomous systems. *IEEE International Symposium on Intelligent Control/Intelligent Systems and Semiotics*. 1999, Cambtridge, MA, 308-313.

9. Stover JA, Hall DL, Gibson RE. A fuzzy logic architecture for autonomous multisensory data fusion. *IEEE Transactions on Industrial Electronics.* 1996, 43: 403-410.

10. Kumar R, Stover JA. A behaviour-based intelligent control architecture with application to coordination of multiple underwater vehicles.  *IEEE Transactions on Systems, Man and Cybernetics.* 2000, 30: 767-784.

11. Laird JE, Newell A, Rosenbloom PS. Soar: An architecture for general intelligence.  *Artificial Intelligence*. 1987, 33(3): 1-64.

12. Anderson JR, Bothell D, Byrne MD, Douglas S, Lebiere C, Qin Y. An integrated theory of mind. *Psychological Review*. 2004, 11(4):1026-1060.

13. Jones RM, Laird JE, Nielsen RE, Coulter KJ, Kenny R, Koss FV. Automated intelligent pilots for combat flight simulations. *AI Magazine*. Spring 1999, 27-41.

14. Bugajska MD, Schultz AC, Trafton JG, Taylor M, Mintz FE. A hybrid cognitive-reactive multi-agent controller. *Proceedings of the 2002 IEEE/RSJ International Conference on Intelligent Robots and Systems*. 2002, Lausanne, 2807-2812.

15. Barbera T, Albus J, Messina E, Schlenoff C, Horst J. How task analysis can be used to derive and organize knowledge for the control of autonomous vehicles. *Robotics and Autonomous Systems*. 2004, 49(1-2): 67-78.

[16] Automated driving: Levels of driving automation are defined in new SAE International Standard J3016. http://www.sae.org/misc/pdfs/automated_driving.pdf (Accessed on 24 August 2018).

[17] Antsaklis PJ, Passino KM, Wang SJ. An introduction to autonomous control systems.  *IEEE Control Systems*. 1991, 11 (4): 5–13.  doi:10.1109/37.88585

[18] Wood SP, Chang J, Healy T, Wood J. The potential regulatory challenges of increasingly autonomous motor vehicles. 2012. *52nd Santa Clara Law Review*. 4 (9): 1423–1502.

[19] Deep learning. Nvidia Accelerated Computing. https://developer.nvidia.com/deep-learning (Accessed 24 August 2018.)

[20] Pal, K. The 5 most amazing AI advances in autonomous driving. April 2018. *Techopedia.* https://www.techopedia.com/the-5-most-amazing-ai-advances-in-autonomous-driving/2/33178 (Accessed 24 August 2018.)

[21] Brandom R. Self driving cars are headed toward an AI roadblock. July 2018. T*he Verge.* https://www.theverge.com/2018/7/3/17530232/self-driving-ai-winter-full-autonomy-waymo-tesla-uber (Accessed 24 August 2018)

[22] Knight W. Finally, a driverless car with some common sense. September 2017. *MIT Technology Review*. https://www.technologyreview.com/s/608871/finally-a-driverless-car-with-some-common-sense/ (Accessed 24 August 2018)

[23] Road safety annual report 2015. OECD/ITF. Paris: International Traffic Safety Data and Analysis Group, International Transport Forum. https://www.oecd-ilibrary.org/transport/road-safety-annual-report-2015_irtad-2015-en (Accessed 24 August 2018)

[24] Sekiguchi K, Tanaka K, Hori K. "Design with discourse" for design from the ethics level. *Proceedings of the 2010 Conference on Information Modelling and Knowledge Bases XXI*. 2010, pp. 307-314.

[25] Spiekermann S. *Ethical IT Innovation: A Value-Based System Design Approach*. 2015. CRC Press.

[26] IEEE. The IEEE Global Initiative on Ethics in Autonomous and Intelligent Systems. 2017. Available from https://standards.ieee.org/develop/indconn/ec/ead_executive_summary_v2.pdf (Accessed 24 August 2018)

[27] Russell SN, Norvig P. The ethics and risks of developing artificial intelligence. 2009 (3rd Edition). In *Artificial Intelligence: A Modern Approach.* Chapter 26.3*.* Prentice Hall.

[28] Bostrom N, Yudkowsky E. The ethics of artificial intelligence. 2014. In K. Frankish, W.M. Ramsey (Eds), *The Cambridge Handbook of Artificial Intelligence.* Chapter 15. CUP.

[29] Anderson M, Anderson SL (Eds). *Machine Ethics.* 2011. CUP.

[30] Arvan M. Mental time-travel, semantic flexibility, and AI ethics. *AI & Society.* 2018. https://doi.org/10.1007/s00146-018-0848-2 (Accessed 24 August 2018).

[31] Anderson SL. Machine metaethics. In Anderson and Anderson (Eds.), *Machine Ethics*. 2011. CUP.

[32]Allen C, Wallach W, Smit I. Why machine ethics? In Anderson and Anderson (Eds.), *Machine Ethics*. 2011. CUP.

[33] Anderson SL. Philosophical concerns with machine ethics. In Anderson and Anderson (Eds.), *Machine Ethics*. 2011. CUP.

[34] Floridi L. On the morality of artificial agents. In Anderson and Anderson (Eds.), *Machine Ethics.* 2011. CUP.

[35] McLaren BM. Computational models of ethical reasoning: Challenges, Initial Steps, and Future Directions. *IEEE Intelligent Systems.* 2006, 21(4): 29-37. Reprinted in Anderson and Anderson (Eds.), *Machine Ethics.* 2011. CUP.

[36] Guarini M. Computational neural modelling and the philosophy of ethics: Reflections on the particularism-generalism debate. In Anderson and Anderson (Eds.), *Machine Ethics.* 2011. CUP.

[37] Mackworth AK. Architectures and ethics for robots: Constraint satisfaction as a unitary design framework. In Anderson and Anderson (Eds.), *Machine Ethics.* 2011. CUP.

[38] Turilli M. Ethical protocols design. In Anderson and Anderson (Eds.), *Machine Ethics.* 2011. CUP.

[39] Bringsjord S, Arkoudas K, Bello, P. Towards a general logicist methodology for engineering ethically correct robots. *IEEE Intelligent Systems.* 2006, 21(4):38-44.

[40] Bringsjord S, Taylor J, van Heuveln B, Arkoudas K, Clark M, Wojtowicz R. Piagetian roboethics via category theory: Moving beyond mere formal operations to engineer robots whose decisions are guaranteed to be ethically correct. In Anderson and Anderson (Eds.), *Machine Ethics.* 2011. CUP.

[41] Pereira LM, Saptawijaya A. Modeling morality with prospective logic. In Anderson and Anderson (Eds.), *Machine Ethics.* 2011. CUP.

[42] Anderson SL, Anderson M. A prima-facie duty approach to machine ethics: Machine learning of features of ethical dilemmas, prima facie duties, and decision principles through a dialogue with ethicists. In Anderson and Anderson (Eds.), *Machine Ethics.* 2011. CUP.

[43] Anderson SL, Anderson M. A prima facie duty approach to machine ethics. In Anderson and Anderson (Eds.), *Machine Ethics.* 2011. CUP.

[44] Garcia M. Racist in the machine: The disturbing implications of algorithmic bias. 2016. *World Policy Journal*, 33(4):111-117. Accessed October 2018, from http://muse.jhu.edu/article/645268/pdf.

[45] Devlin H. AI programs exhibit racial and gender biases, research reveals. Accessed October 2018, from https://www.theguardian.com/technology/2017/apr/13/ai-programs-exhibit-racist-and-sexist-biases-research-reveals.

[46] Fuchs DJ. The dangers of human-like bias in machine-learning algorithms. *Missouri S&T's Peer to Peer*, 2(1). Accessed October 2018, from http://scholarsmine.mst.edu/peer2peer/vol2/iss1/1.

[47] Zadeh LA. Fuzzy sets. *Information and Control*. 1996, 8(3):338-353. doi:10.1016/S0019-9958(65)90241-X

[48] Novak V. Fuzzy logic, fuzzy sets, and natural languages. *International Journal of General System*. 1991, 20 (1): 83-97. doi: 10.1080/03081079108945017

[49] Van Leekwijck Q, Kerre EE. Defuzzification: Criteria and classification. *Fuzzy Sets and Systems*.1999, 108: 159-178. doi:10.1016/S0165-0114(97)00337-0

[50] Martin B. Instance-Based Learning: Nearest Neighbour with Generalisation. 1995. Working Paper Series 95/18. Hamilton, New Zealand: University of Waikato, Department of Computer Science. https://researchcommons.waikato.ac.nz/handle/10289/1095. Accessed in October 2018.

[51] Wettschereck D. and Dietterich T.G. An Experimental Comparison of the Nearest-Neighbor and Nearest-Hyperrectangle Algorithms. *Machine Learning*. 1995, 19: 1–25.

[52] Gerla G. Vagueness and formal fuzzy logic: Some criticisms. *Logic and Logical Philosophy*. 2017, 26: 431-460. http://dx.doi.org/10.12775/LLP.2017.031 . Accessed in November 2018.

[53] Syropoulos A. *Theory of Fuzzy Computation.* IFSR International Series on Systems Science and Engineering. New York, NY, Springer, vol. 31, 2014.

[54] The IEEE Global Initiative on Ethics of Autonomous and Intelligent Systems. *Ethically Aligned Design: A Vision for Prioritizing Human Well-being with Autonomous and Intelligent Systems*, Version 2. IEEE, 2017. http://standards.ieee.org/develop/indconn/ec/autonomous_systems.html.

# Supplementary Information (SI)

*SI, Table 1: Fuzzy models, variables, membership functions and MATLAB parameters for an IAV to decide when to take over control from a human driver*

| Fuzzy moral dimension (model) | Variable | Range | Membership functions | Shape | Parameters |
|---|---|---|---|---|---|
| Right_wrong (3 input, 1 output), centroid | Distance (input) | 1-10 | 2 (low risk; high risk) | trapezoidal | [0, 0, 5, 6]; [5, 6, 10, 10] |
| | Lane: input | 1-10 | 2 (low risk; high risk) | trapezoidal | [0, 0, 8, 9]; [7, 8, 10, 10] |
| | Speed: input | 1-100 | 2 (low risk; high risk) | trapezoidal | [0, 0, 40, 80]; [40, 80, 100, 100] |
| | Take control right or wrong (tcrightwrong): output | 1-10 | 2 (tcwrong; tcright) | zmf; smf | [7, 10]; [4, 7] |
| Good_bad (3 input, 1 output), centroid | Distance: input | 1-10 | 2 (low risk; high risk) | trapezoidal | [0, 0, 5, 6]; [5, 6, 10, 10] |
| | Lane: input | 1-10 | 2 (low risk; high risk) | trapezoidal | [0, 0, 8, 9]; [7, 8, 10, 10] |
| | Speed: input | 1-100 | 2 (low risk; high risk) | trapezoidal | [0, 0, 40, 80]; [40, 80, 100, 10] |
| | Take control good or bad (tcgoodbad): output | 1-10 | 2 (tcbad; tcgood) | zmf; smf | [4,8]; [2,6] |
| Virtuous meta ethics controller (2 input, one output, centroid | Take control right or wrong (tcrw): input | 1-10 | 2 (rwdtc; rwtc) | zmf; smf | [1, 10]; [1,10] |
| | Take control good or bad (tcgb) | 1-10 | 2 (gbdtc; gbtc) | trapezoidal | [0, 0, 5, 6]; [5, 6, 10, 10] |
| | Control | 1-10 | 2 | zmf; smf | [5.5, 10]; [1, 5.5] |

*SI, Table 2: Parameters for variables and membership functions of the moral dilemma reasoner in MATLAB (Figure 8). Note that 'gbelmf' stands for a bell-shaped membership function and 'gaussmf' for a Gaussian-shaped membership function.*

| Fuzzy dilemma reasoner | Variable | Range | Membership functions | Shape | Parameters |
|---|---|---|---|---|---|
| **Dilemma_right_wrong (2 input, 1 output), centroid** | Straight: input | 1-10 | 2 (low risk of death; high risk of death) | trapezoidal | [0, 0, 2,3]; [2,3,10,10] |
| | Swerve: input | 1-10 | 2 (low risk of death; high risk of death) | trapezoidal | [0, 0, 2,3]; [2,3,10,10] |
| | Rightwrong: output | 1-10 | 2 (swerve is wrong; swerve is right) | zmf;smf | [5,6]; [5,6] |
| **Dilemma_good_bad (1 input, 1 output), centroid** | Pedestrian: input | 1-10 (age *0.1) | 2 (young; notyoung) | trapezoidal | [0, 0, 2,3]; [2,5,10,10] |
| | Avoid: output | 1-10 | 2 (avoid is bad; avoid is good) | zmf;smf | [4,8]; [2,6] |
| **Dilemma_controller (2 input, one output, centroid)** | Deathrisk: input | 1-10 | 2 (deathrisklow; deathriskhigh) | gbelmf | [4.5, 3, 1]; [4.5, 2.5, 10] |
| | Pedestrianrisk: input | 1-10 | 2 (avoidbad; avoidgood) | gbelmf | [5, 2.5, 0]; [5, 2.5, 10] |
| | Dilemmadecision: output | 1-10 | 2 (straightahead; swerve) | gaussmf | [1,3]; [1,7] |